\documentclass[twoside,11pt]{article}
\usepackage{jmlr2e}
\usepackage[utf8]{inputenc}
\usepackage{amsmath}
\usepackage{amsfonts}
\usepackage{amssymb}
\usepackage{tabularx}
\usepackage{graphicx}
\usepackage{booktabs}
\usepackage{textcomp}

\usepackage[labelfont=bf]{caption}

\ShortHeadings{AutoML vs.\ Deep Learning}{Romano \textit{et al}}
\firstpageno{1}

\begin{document}

\title{Is deep learning necessary for simple classification tasks?}

\author{%
  \name Joseph~D.~Romano \email joseph.romano@pennmedicine.upenn.edu \\
  \addr Institute for Biomedical Informatics\\
  University of Pennsylvania\\
  Philadelphia, PA 19104, USA
  \AND
  \name Trang~T.~Le
  \email ttle@pennmedicine.upenn.edu\\
  \addr Institute for Biomedical Informatics\\
  University of Pennsylvania\\
  Philadelphia, PA 19104, USA
  \AND
  \name Weixuan~Fu
  \email weixuanf@pennmedicine.upenn.edu\\
  \addr Institute for Biomedical Informatics\\
  University of Pennsylvania\\
  Philadelphia, PA 19104, USA
  \AND
  \name Jason~H.~Moore
  \email jhmoore@upenn.edu\\
  \addr Institute for Biomedical Informatics\\
  University of Pennsylvania\\
  Philadelphia, PA 19104, USA
}

\editor{N/A}

\maketitle
\begin{abstract}
  Automated machine learning (AutoML) and deep learning (DL) are two
  cutting-edge paradigms used to solve a myriad of inductive learning
  tasks.  In spite of their successes, little guidance exists for when
  to choose one approach over the other in the context of specific
  real-world problems.  Furthermore, relatively few tools exist that
  allow the integration of both AutoML and DL in the same analysis to
  yield results combining both of their strengths.  Here, we seek to
  address both of these issues, by (1.) providing a head-to-head
  comparison of AutoML and DL in the context of binary classification
  on 6 well-characterized public datasets, and (2.) evaluating a new
  tool for genetic programming-based AutoML that incorporates deep
  estimators.  Our observations suggest that AutoML outperforms simple
  DL classifiers when trained on similar datasets for binary
  classification but integrating DL into AutoML improves
  classification performance even further.  However, the substantial
  time needed to train AutoML+DL pipelines will likely outweigh
  performance advantages in many applications.
\end{abstract}

\section{Introduction and Background}
Deep learning (DL) and automated machine learning (AutoML) are two
approaches for constructing high-performance estimators that
dramatically outperform traditional machine learning in a variety of
scenarios, especially on classification and regression tasks. In spite
of their successes, there remains substantial debate---and little
quantitative evidence---on the practical advantages of the two
approaches and how to determine which will perform best on specific
real-world problems.

To address this issue, we conduct a series of experiments to compare
the performance of DL and AutoML pipelines on 6 well-characterized
binary classification problems.  We also introduce and critically
assess a new resource that leverages the strengths of both approaches
by integrating deep estimators into an existing AutoML tool for
constructing machine learning pipelines using evolutionary
algorithms. Specifically, we sought to answer two questions in this
study:
\begin{enumerate}
\item How well does genetic programming-based AutoML perform on simple
  binary classification tasks in comparison to DL?
\item By augmenting AutoML with DL estimators, do we achieve better
  performance than with either of the two alone?
\end{enumerate}

Beyond these two questions, we also provide specific recommendations
on choosing between DL and AutoML for simple classification tasks, and
a set of priorities for future work on the development of methods that
integrate DL and AutoML. Our new software resource---which we have
named TPOT-NN---is an extension to the previously described Tree-based
Pipeline Optimization Tool (TPOT), and is freely available online.

\subsection{Deep learning and artificial neural networks}
Deep learning is an umbrella term used to describe statistical
methods---usually involving artificial neural networks (ANNs)---that
provide estimators consisting of multiple stacked
transformations~\citep{schmidhuber2015deep,lecun2015deep}. Traditional
estimators (i.e., ``shallow'' models, such as logistic regression or
random forest classifiers) are limited in their ability to approximate
nonlinear or otherwise complex objective functions, resulting in
reduced performance, particularly on datasets where entities with
similar characteristics are not cleanly separated by linear decision
boundaries. Meanwhile, DL estimators overcome this limitation through
sequential nonlinearities. It follows that, given a sufficient number
of layers and an appropriate learning algorithm, even structurally
simple feed-forward neural networks with a finite number of neurons
can approximate virtually any continuous function on compact subsets
of Euclidean space~\citep{leshno1993multilayer}.

Nonetheless, the successes of DL have been tempered by a number of
important criticisms.  Compared to shallow models, it is substantially
more complex to parameterize a deep ANN due to the explosion of free
parameters that results from increased depth of the network or width
of individual layers~\citep{shi2016benchmarking}.  Furthermore, DL
models are notoriously challenging to interpret since the features in
a network's intermediate layers are a combination of features from all
previous layers, which effectively obscures the intuitive meaning of
individual feature weights in most nontrivial
cases~\citep{lipton2018mythos,lou2012intelligible}.

It is also worth noting that DL model architectures can reach immense
sizes. For example, the popular image classification network ResNet
performed best in an early study when constructed with 110
convolutional layers, containing over 1.7 billion tunable
parameters~\citep{he2016deep}.  However, for standard binary
classification on simple datasets, smaller DL architectures can still
substantially outperform shallow learners, both in terms of error and
training time~\citep{auer2002reducing,collobert2004links}.  For the
purpose of establishing a baseline comparison between AutoML and DL,
we restrict our analyses in this study to this latter case of
applications.

\subsection{Automated Machine Learning}
One of the most challenging aspects of designing an ML system is
identifying the appropriate feature transformation, model architecture
and hyperparameterization for the task at hand.  For example, count
data may benefit from a square-root transformation.  Similarly, a
support-vector machine (SVM) model might predict more accurately
susceptibility to a certain complex genetic disease than a gradient
boosting model trained on the same dataset.  Further, different
choices of hyperparameters within that SVM model of kernel function
$k$ and soft margin width $C$ can lead to wildly different
performances.  Traditionally, these architecture considerations need
to be made with the help of prior experience, brute-force search, or
experimenter intuition, all of which are undesirable for their own
reasons.

AutoML, on the other hand, provides methods for automatically
selecting these options from a universe of possible architecture
configurations. A number of different AutoML techniques can be used to
find the best architecture for a given task, but one that we will
focus on is based on \textit{genetic programming} (GP). Broadly, GP
constructs trees of mathematical functions that are optimized with
respect to a fitness metric such as classification accuracy
~\citep{banzhaf1998genetic}.  Each generation of trees is constructed
via random mutations to the tree's structure or the operations
performed at each node in the tree.  Repeating this process for a
number of training generations produces an optimal tree.  Like in
natural evolution, increasingly more fit architectures are propagated
forward while less fit architectures ``die out''.

TPOT (Tree-based Pipeline Automation Tool) is a Python-based AutoML
tool that uses genetic programming to identify optimal ML pipelines
for either regression or classification on a given (labeled)
dataset. Briefly, TPOT performs GP on trees where nodes are comprised
of \textit{operators}, each of which falls into one of four operator
types: preprocessors, decomposition functions, feature selectors, or
estimators (i.e., classifiers and regressors). Input data enters the
tree at leaf nodes, and predictions are output at the root node. Each
operator has a number of free parameters that are optimized during the
training process. TPOT maintains a balance between high performance
and low model complexity using the NSGA-II Pareto optimization
algorithm~\citep{deb2002fast}. As a result, the pipelines learned by
TPOT consist of a relatively small number of operators (e.g., in the
single-digits) that can still meet or exceed the performance of
competing state-of-the-art ML approaches.

In theory, TPOT can construct pipelines that are structurally
equivalent to DL estimators by stacking multiple shallow estimators in
a serial configuration. However, since the individual objectives for
each operator are decoupled, it is unclear whether these
configurations can attain the same performance as standalone DL
estimators, such as multilayer perceptrons (MLPs).

With the exception of our new deep learning estimators (which are
implemented in PyTorch), all operators are implemented in either
Scikit-learn~\citep{scikit-learn} or XGBoost~\citep{chen2016xgboost},
both of which are open-source, popular Python-based machine learning
libraries. TPOT natively performs cross validation and generates
Python scripts that implement the learned pipelines. For a more
detailed description and evaluations of TPOT, please
see~\citep{OlsonGECCO2016, Olson2016EvoBio, le2020scaling}.

\begin{figure}[h]
  \centering
  \includegraphics[width=\textwidth]{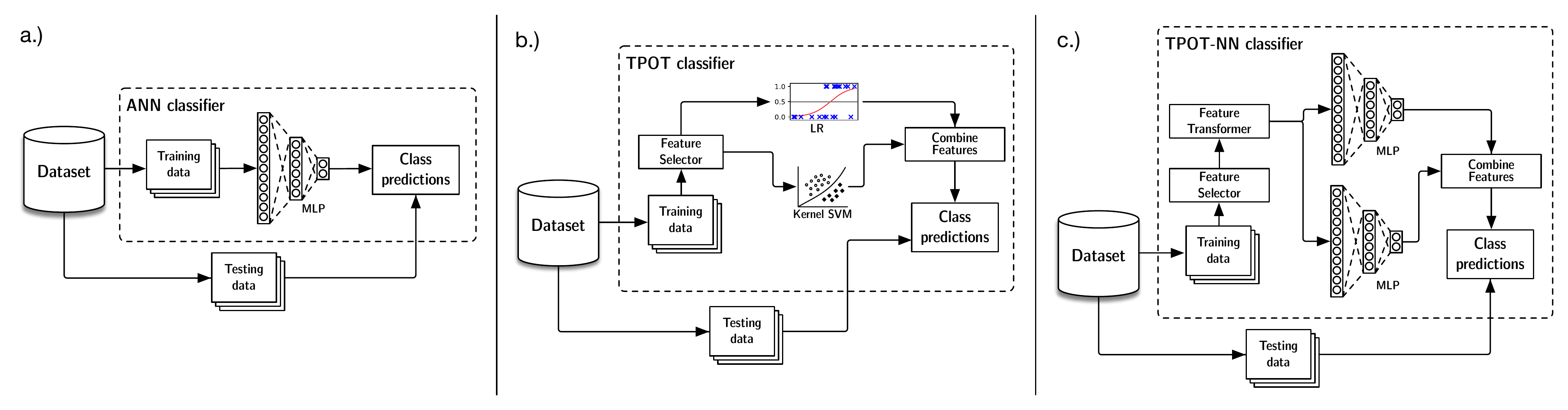}
  \caption{Example pipelines for model configurations used in this
study. \textbf{a.)} Deep learning strategy with no
AutoML. \textbf{b.)} Example of a standard (no neural networks) TPOT
pipeline containing a logistic regression classifier and a kernel SVM
classifier. \textbf{c.)} Example of a TPOT-NN pipeline containing two
multilayer perceptron estimators.}\label{fig:abstract}
\end{figure}

\section{Results}
We evaluated the pipelines' performance based on two metrics:
classification accuracy of the trained pipeline and the elapsed time
to train the pipeline.  Each experiment can be grouped into 1 of 3
main configurations: NN (a single MLP classifier with no GP); TPOT
(pipelines learned using GP, possibly containing multiple stacked
estimators, preprocessors, and feature transformers); and TPOT-NN (the
same as TPOT, but with added MLP and PyTorch logistic regression
estimators). The contrived examples in
\textbf{Figure~\ref{fig:abstract}} illustrate the differences between
pipelines learned from the 3 configurations.

In \textbf{\S\ref{baseline-estimators}}, we describe MLP architectures
in this study as well as rationale for our specific design decisions.

\subsection{Model performance comparison---NN, TPOT, and TPOT-NN}

\begin{figure}[h]
  \centering
  \includegraphics[width=\textwidth]{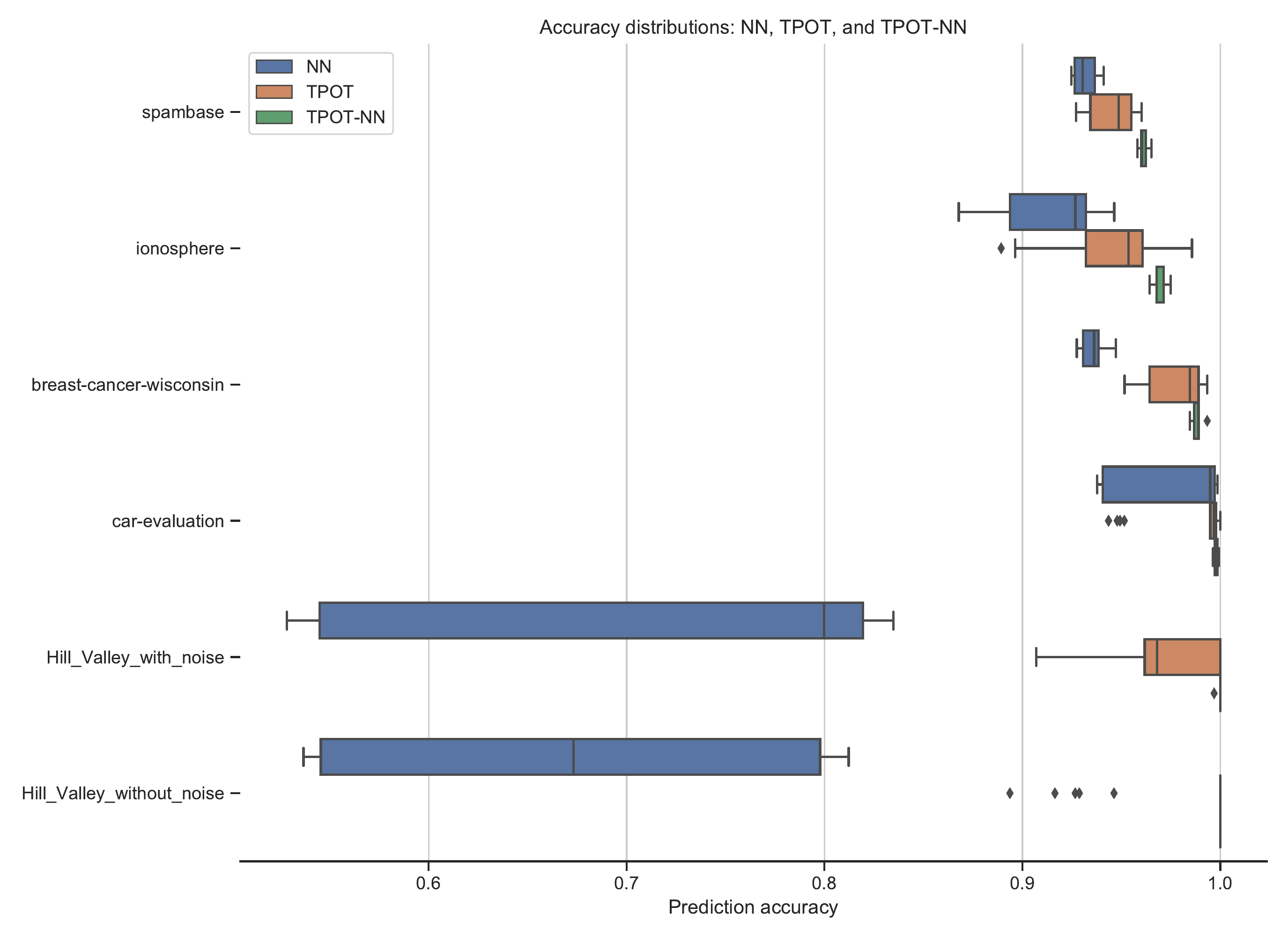}
  \caption{Prediction accuracy of neural network classifiers (NN),
    TPOT classifiers, and TPOT-NN classifiers tested on 6 open-access
    datasets. Each point used to construct the distributions
    corresponds to the accuracy of one trained
    pipeline.}\label{fig:config}
\end{figure}

The prediction accuracy distributions of our experiments are shown in
\textbf{Figure~\ref{fig:config}}. For each of the 6 datasets, neural
network estimators alone yielded the lowest average prediction
accuracy while the TPOT-NN pipelines performed best. In general, the
TPOT-NN pipelines performed only marginally better than the standard
TPOT pipelines. Notably, the (non-TPOT) neural network approach
yielded substantially poorer performance on two of the datasets
(\texttt{Hill\_Valley\_with\_noise} and
\texttt{Hill\_Valley\_without\_noise}). We discuss a likely
explanation in~\textbf{\S\ref{consistency}}.

\subsection{Validating TPOT-NN's neural network estimators}

\begin{figure}[h]
  \centering
  \includegraphics[width=1.1\textwidth]{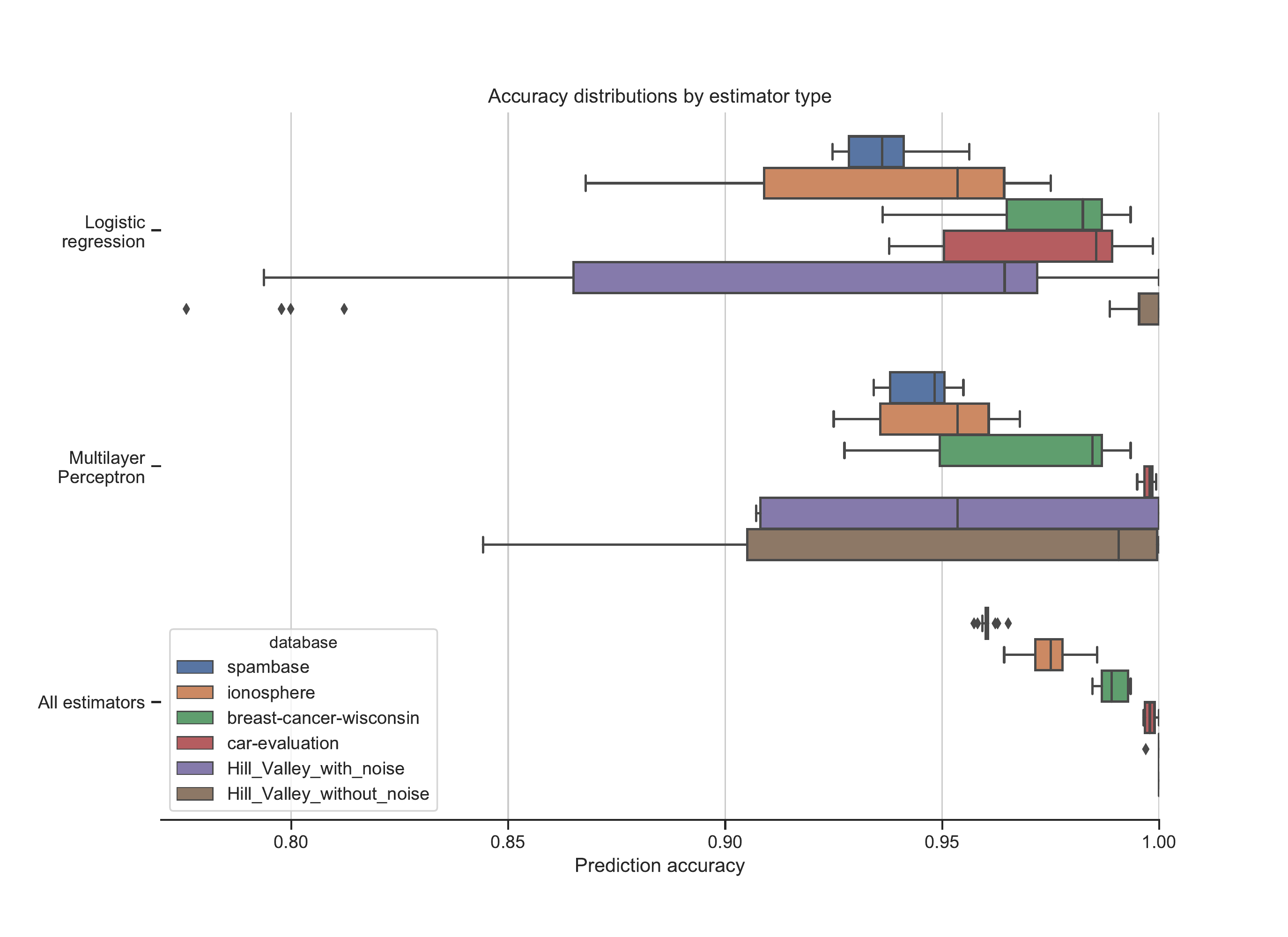}
  \caption{Prediction accuracy on each of the 6 evaluation databases
    stratified by estimators enabled in TPOT. Each point used to
    construct the box plots is the classification accuracy of a single
    optimized TPOT pipeline.}\label{fig:lr}
\end{figure}

The performance advantages of DL estimators are largely derived from
their ability to fit complex nonlinear objectives, which is a
consequence of stacking multiple neural layers in a serial
configuration. To confirm that the MLP estimator included in TPOT-NN
leverages this advantage, we used the TPOT-NN API to design a logistic
regression (LR) classifier, which is functionally equivalent to the
MLP classifier with the intermediate (`hidden') layer removed---a
shallow estimator implemented identically to the TPOT-NN MLP. We then
ran a series of experiments testing each of these classifiers alone
(i.e., only using TPOT to optimize model hyperparameters, and not to
construct pipelines consisting of multiple operators). The results of
these experiments are summarized in \textbf{Figure~\ref{fig:lr}}. As a
means for comparison, we visualized these results alongside
experiments using the full TPOT-NN implementation, to confirm that the
inclusion of all TPOT operators results in further improvements to
classification accuracy.

The results of these experiments support this intuition: TPOT-NN's
implementation of MLP classifiers performs better than the identically
implemented LR classifier in most cases. Enabling all operators
available to TPOT-NN results in an even more dramatic performance
increase in addition to a remarkable improvement in the consistency of
results on individual datasets. Together, these observations support
the claims that (a.) TPOT-NN's new estimators leverage the increased
estimation power characteristic of deep neural network models, and
(b.) TPOT-NN provides an effective solution to improve on the results
yielded by ``na\"ive'' DL models on simple classification tasks.

\subsection{Training duration of TPOT-NN models}\label{train-time}
\begin{figure}[h]
    \centering
    \includegraphics[width=0.9\textwidth]{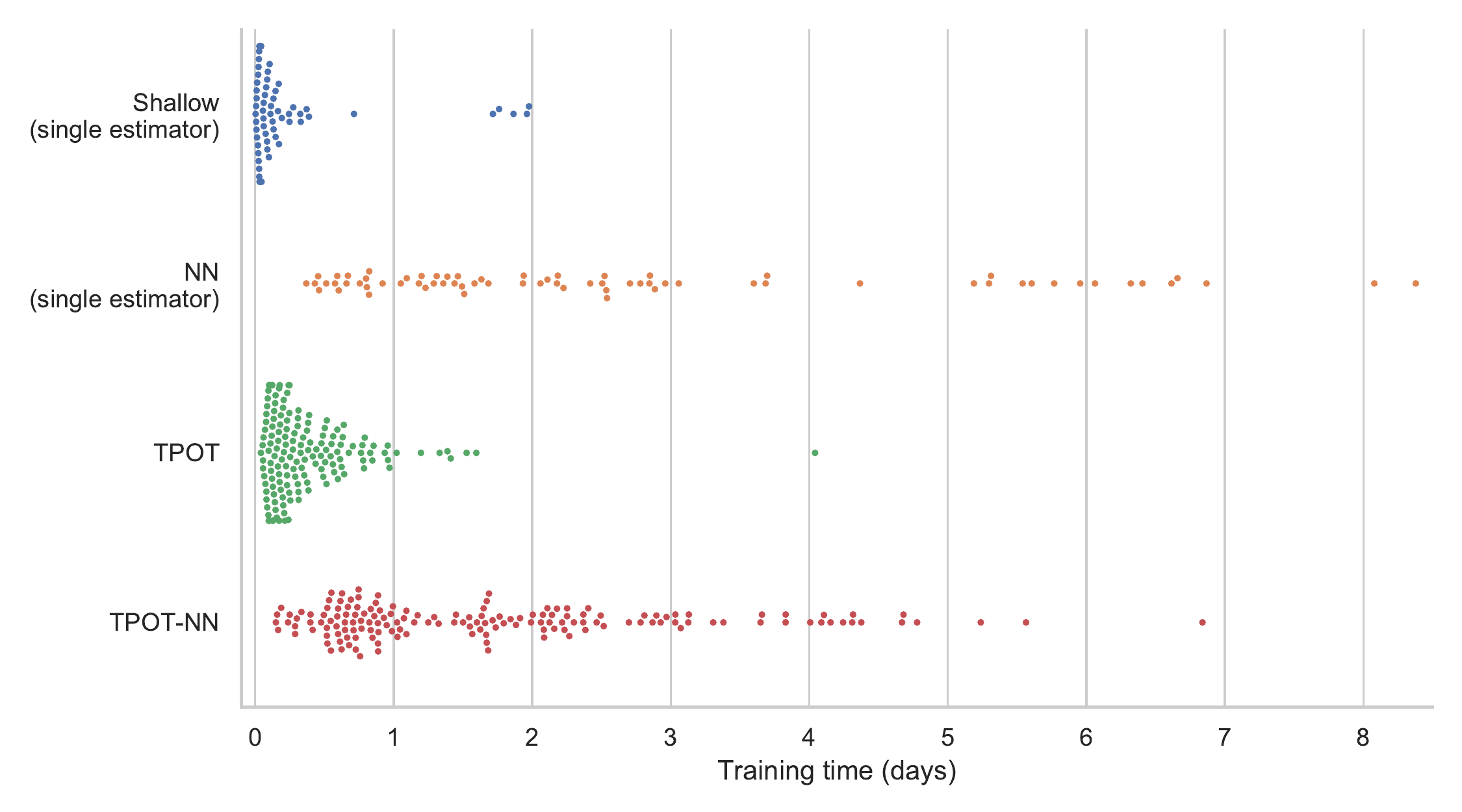}
    \caption{Comparison of shallow estimator, NN estimator (MLP),
      TPOT, and TPOT-NN experiment training time distributions.}
    \label{fig:time}
\end{figure}

The total training time for TPOT pipelines ranged from 4h~22m to
8d~19h~55m, with a mean training time of 1d~0h~49m ($\pm$
1d~13h~28m). \textbf{Table~\ref{tab:time}} and
\textbf{Figure~\ref{fig:time}} show how training times vary across the
different experiments. To provide a basis of comparison for standard
TPOT, we include training time for pipelines consisting of a single
shallow estimator only, referred to as ``Shallow (single estimator)'',
where TPOT uses GP solely for the purpose of optimizing
hyperparameters on a pipeline containing a single estimator. Both
configurations involving NN estimators required a substantially larger
amount of time to train a single pipeline on average: an increase of
629\% for NN pipelines and 336\% for TPOT-NN piplelines versus
standard TPOT. Future studies on larger datasets should be performed
to establish how this relationship scales with respect to training set
size.

\begin{table}[h]
  \centering
  \begin{tabular}{l r r r r}
    \toprule
    Experiment type&Mean training time&Minimum&Maximum&Std. dev.\\
    \midrule
    Shallow (single estimator)&5h 39m&4m&1d 23h 26m&11h 30m\\
    TPOT&0d 9h 26m&59m&4d 59m&10h 23m\\
    NN (single estimator)&2d 20h 48m&8h 51m&8d 19h 55m&2d 7h 19m\\
    TPOT-NN&1d 17h 8m&3h 35m&6d 20h 6m&1d 7h 2m\\
    \bottomrule
  \end{tabular}
  \caption{Training time statistics for single shallow estimators,
    `standard' TPOT, single neural network estimators, and TPOT-NN. In
    all cases, GP is enabled for the purpose of tuning
    hyperparameters.}
  \label{tab:time}
\end{table}

\subsection{Structural topologies of pipelines learned by GP}\label{topologies}
TPOT assembles pipelines that consist of multiple operators---possibly
including multiple classifiers or regressors in addition to feature
selectors and feature transformers---to achieve better performance
than individual machine learning
estimators~\citep{OlsonGECCO2016}. Since the estimation capacity of
simple feedforward neural networks is a monotonic function of network
depth, we sought to determine whether TPOT can automatically construct
deep architectures by stacking shallow estimators in the absence of
\textit{a priori} instruction to do so.

When TPOT-NN was forced to build pipelines comprised only of feature
selectors, feature transformers, and logistic regression estimators,
it did indeed construct pipelines consisting of stacked arrangements
of logistic layers that strongly resemble well-known DL models. The
following Python code is the output of one of these, selected at
random from the pool of LR-only TPOT-NN pipelines (hyperparameters
have been removed for readability):

% File name: tpot-nn_lr_nn_solo_config_carevaluation_rep4_1585579789.py
\begin{verbatim}
# Average CV score on the training set was: 0.9406477266781772
exported_pipeline = make_pipeline(
    make_union(
        StackingEstimator(estimator=make_pipeline(
            StackingEstimator(estimator=PytorchLRClassifier(...)),  # LR1
            StackingEstimator(estimator=PytorchLRClassifier(...)),  # LR2
            PytorchLRClassifier(...)  # LR3
        )),
        FunctionTransformer(copy)  # Identity (skip)
    ),
    PytorchLRClassifier(...) # LR4
)
\end{verbatim}

\begin{figure}[h]
    \centering
    \includegraphics[width=\textwidth]{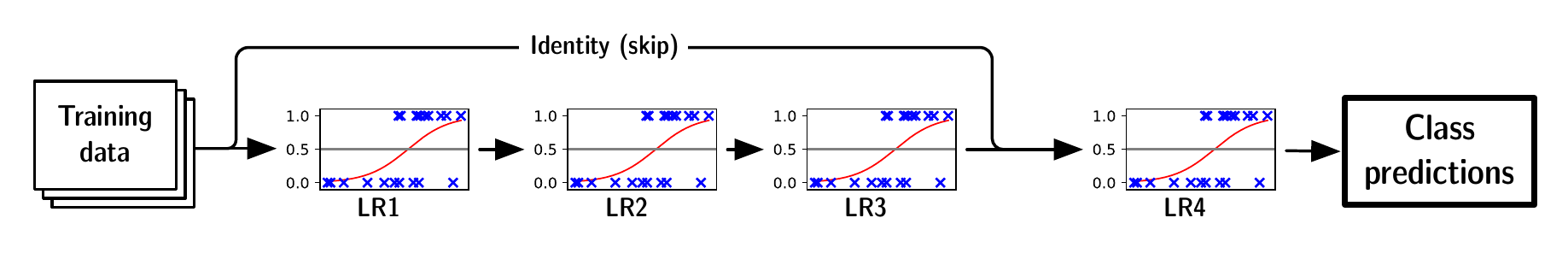}
    \caption{Randomly selected pipeline learned when restricting
      TPOT's pool of estimators to logistic regression classifiers
      only. The structure of the pipeline resembles a residual
      block---a major component of the ResNet DL architecture.}
    \label{fig:res-block}
\end{figure}

The structure of this pipeline is virtually identical to a residual
block---one of the major innovations that has led to the success of
the ResNet architecture. A graphical representation of this pipeline
is shown in \textbf{Figure~\ref{fig:res-block}}. This suggests that
AutoML could be used as a tool for identifying new submodules for
larger DL models. We discuss this possibility further
in~\textbf{\S\ref{discovery}}.

\section{Discussion}
\subsection{Assessing the tradeoff between model performance and
  training time}
The amount of time needed to train a pipeline is an important
pragmatic consideration in real-world applications of ML. This
certainly extends to the case of AutoML: The parameters we use for
TPOT include 100 training generations with a population size of 100 in
each generation, meaning that we effectively evaluate 10,000
pipelines---each of which consists of a variable number of
independently optimizable operators---for every experiment (of which
there were 1,375 in the present study). As demonstrated
in~\textbf{\S\ref{train-time}}, we generally expect a pipeline to
train (in the case of `standard' TPOT) in the range of several hours
to slightly over 1 day.

Our relatively simple MLP implementation (see
\textbf{\S\ref{baseline-estimators}}) sits at the lower
end---complexity wise---of available deep learning estimators, and
likewise is one of the fastest to train. Regardless, including the MLP
estimator in a TPOT experiment increases the average time to learn a
pipeline by at least 4-fold. Users will have to determine, on an
individual basis and dependent on the use case, whether the potential
accuracy increase of at most several percentage points is worth the
additional time and computational investment.

Nonetheless, the data in \textbf{Figure~\ref{fig:config}} demonstrate
that it is unlikely for a TPOT-NN pipeline to perform \textit{worse}
than a (non-NN) TPOT pipeline. In `mission critical' settings where
training time is not a major concern, TPOT-NN can be expected to
perform \textit{at least} as well as standard TPOT.

\subsection{AutoML is effective at recovering DL performance
  consistency}\label{consistency}
One of the most striking patterns in our results is that the NN-only
models (both MLP and LR) yielded highly inconsistent performance on
several datasets, especially the two ``hill/valley'' datasets (see
\textbf{Figure~\ref{fig:config}}).  However, this is unsurprising:
these datasets contain sequence data, where an estimator must be able
to identify a `hill' or a `valley' that could occur at any location in
the sequence of 100 features.  Therefore, the decision boundary of an
estimator over these data is being optimized over a highly non-convex
objective function.  Standard DL optimization algorithms struggle with
problems like these while heuristic and ensemble methods tend to
perform well more consistently, explaining the large difference in
classification accuracy variance between NN-only and TPOT-NN
experiments.

\subsection{AutoML as a tool to discover novel DL architectures}\label{discovery}
Based on the results we describe in~\textbf{\S\ref{topologies}},
AutoML (and TPOT-NN, in particular) may be useful for discovering new
neural network ``motifs'' to be composed into larger networks. For
example, by repeating the internal architecture shown in
\textbf{Figure~\ref{fig:res-block}} to a final depth of 152 hidden
layers and adjusting the number of nodes in those layers, the result
is virtually identical to the version of ResNet that won 1st place in
5 categories at two major image recognition competitions in
2015~\citep{he2016deep}. In the near future, we plan to investigate
whether this phenomenon could be scaled into a larger, fully
data-driven approach for generating modular neural network motifs that
can be composed into models effective for a myriad of learning tasks.

\subsection{Future work on integrating AutoML and DL}
Since one of our primary goals in this work was to provide a baseline
for future development of deep learning models in the context of
AutoML, the two PyTorch models we have currently built (logistic
regression and MLP) are structurally simple. Future work on TPOT-NN
will allow expansion of its functionality to improve the capabilities
of the existing models as well as incorporate other, more complex
architectures, such as convolutional neural networks, recurrent neural
networks, and deep learning regressors.

\section{Conclusions}
AutoML and DL are immensely useful tools for approaching a wide
variety of inductive learning tasks, and it is clear that both hold
strengths and weaknesses for specific use cases. Rather than viewing
them as \textit{competing} methods, we instead propose that the two
can work synergistically: For at least the cases we explored in this
study (classification on 6 well-characterized datasets with relatively
simple feature correlations), the addition of multilayer perceptron
classifiers into the pool of available operators improves the
performance of AutoML. Since such learned pipelines often explicitly
include feature selection and feature transformation operators, they
provide a feasible mechanism for improving interpretability of models
that make use of DL.

Currently, use of these DL estimators in TPOT significantly increases
training time for pipelines, which likely will limit their
applications in many situations. Nonetheless, this suggests a
multitude of novel directions for methodological research in machine
learning and artificial intelligence. TPOT-NN serves as both an early
case study as well as a platform to facilitate DL+AutoML research in a
reproducible, transparent manner that is open to the scientific
community.

\section{Methods}

\subsection{Datasets and experimental setup}
We evaluated TPOT-NN on 6 well-studied publicly available datasets
that were used previously to evaluate TPOT's base implementation
(i.e., no ANN estimators), shown
in~\textbf{Table~\ref{tab:datasets}}. All datasets are contained in
the PMLB Python package. \texttt{Hill\_Valley\_with\_noise} and
\texttt{Hill\_Valley\_without\_noise} consist of synthetic data; the
rest are real data.  The number of data points, data types for each
feature (i.e., binary, integer, or floating-point decimal), number of
features, and number of target classes are variable across the 6
datasets.

\begin{table}[h]
  \centering
  \begin{tabular}{l  l l}
    \toprule
    \textbf{Parameter}&\textbf{Options}&\textbf{Description}\\
    \midrule
    TPOT-NN enabled?&True&PyTorch LR and MLP estimators are enabled\\[4pt]
    &False&Only non-NN estimators are enabled\\[8pt]
    Template pipeline?&True&Pipelines are restricted to the format:\\
    &&Selector$\rightarrow$Transformer$\rightarrow$Classifier\\[4pt]
    &False&Pipeline structure is learned via GP\\[8pt]
    Included estimator(s)&All&All estimators can be used in pipelines\\[4pt]
    &LR&Only logistic regression can be used\\[4pt]
    &MLP&Only multilayer perceptron can be used\\[8pt]
    TPOT enabled?&True&(Default)\\[4pt]
    &False&GP only used for optimizing hyperparameters\\
    &&(pipelines consist of a single estimator)\\
    \bottomrule 
  \end{tabular}
  \caption{Configuration options used to construct the TPOT/TPOT-NN
    experiments in this study. The ``included estimator'' and ``TPOT
    enabled'' parameters were used for testing and validation of the
    TPOT-NN models.}
  \label{tab:experiments}
\end{table}

We performed 720 TPOT experiments in total, corresponding to all
combinations of the configuration parameters shown in
\textbf{Table~\ref{tab:experiments}}. All configurations were run with
5 replicates on each of the 6 datasets listed in
\textbf{Table~\ref{tab:datasets}}, resulting in 30 experiments per
configuration. We used an 80\%/20\% train/test split on the datasets
and scored pipelines based on classification accuracy with 5-fold
cross-validation.

\begin{table}[h]
  \centering
  \caption{6 datasets used to evaluate TPOT-NN. Names are identical to
    corresponding labels used to denote the dataset in the PMLB Python
    library.}\label{tab:datasets}
  \begin{tabular}{lrrlrr}
    \toprule
    Dataset name&$n$&Features&Data type(s)&Classes&Real/synthetic\\
    \midrule
    \texttt{Hill\_Valley\_with\_noise}&1212&100&Float&2&synthetic\\
    \texttt{Hill\_Valley\_without\_noise}&1212&100&Float&2&synthetic\\
    \texttt{breast-cancer-wisconsin}&569&30&Float&2&real\\
    \texttt{car-evaluation}&1728&21&One-hot&4&real\\
    \texttt{ionosphere}&351&34&Mixed&2&real\\
    \texttt{spambase}&4601&57&Mixed&2&real\\
    \bottomrule
  \end{tabular}
\end{table}

\subsection{Baseline TPOT and NN estimators}\label{baseline-estimators}
For baseline (non-NN) TPOT experiments, we included all default
operators from Scikit-learn and XGBoost, and the default
configurations for all trainable model parameters.

We constructed logistic regression (LR) and multilayer perceptron
(MLP) models in PyTorch to serve as neural network models. LR and MLP
are largely considered the two simplest neural network architectures,
and are therefore suitable for initial evaluation of new machine
learning tools based on ANNs. Since a Scikit-learn LR model is
included in standard TPOT, we are able to directly compare the two LR
implementations to validate that the PyTorch models are compatible
with the TPOT framework, and to quantify the performance variation due
to differences in the internal implementations of equivalent
models. To allow for a similar comparison for MLP, we merged
Scikit-learn's MLP model into TPOT, which has been omitted from the
allowable operators in the past due to lengthy training times and
inconsistent performance compared to MLPs constructed using dedicated
deep learning libraries (such as PyTorch).

\subsection{TPOT-NN}
TPOT users can control the set of available operators---as well as the
trainable parameters and the values they can assume---by providing a
`configuration dictionary' (a default configuration dictionary is used
if the user does not provide one).  We coded the new NN estimators for
TPOT within the main TPOT codebase, but provided a separate
configuration dictionary that includes the NN estimators along with
all default operators.  We wrote the new TPOT-NN models in PyTorch,
but the TPOT-NN API could be adapted to other neural computing
frameworks.  Since TPOT requires that all estimators implement an
identical interface (compatible with Scikit-learn conventions for
fitting a model and transforming data with the fit model), we wrapped
the PyTorch models in classes that implement the necessary methods.

Users can also direct TPOT to utilize the NN models by providing a
`template string' instead of a configuration dictionary.  A generic
template string for MLP might look like
``\texttt{Selector-Transformer-PytorchMLPClassifier}'', instructing
TPOT to fit a series of 3 operators (and their associated
hyperparameters): any feature selector, followed by any feature
transformer, followed by an instance of the PyTorch MLP model.  When
no template string is used, TPOT has the ability to learn pipelines
with more complex structures.

\subsection{Hardware and high-performance computing environment}
All experiments were run on a high-performance computing (HPC) cluster
at the University of Pennsylvania. Each experiment was run on a
compute node with 48 available CPU cores and 256 GB of RAM. Job
scheduling was managed using IBM's Platform Load Sharing Facility
(LSF). All experiments involving PyTorch neural network estimators
were run on nodes equipped with
NVIDIA\textsuperscript{\textregistered} TITAN GPUs.

\section{Code and Data Availability}
TPOT-NN is a submodule included with the full TPOT Python
distribution, which is freely available on the Python Package Index
(PyPI) and through GitHub [REF]. Due to the substantially increased
training time of the neural network models, users must explicitly
enable the use of the TPOT-NN estimators by passing the parameter
\texttt{config='TPOT NN'} when instantiating a TPOT pipeline. The code
we used to evaluate TPOT-NN is available on GitHub in a separate
repository [REF]. A frozen copy of all code, data, runtime output, and
trained models is also available on FigShare [REF].

\section{Acknowledgements}
This work was made possible through NIH grants \texttt{T32-ES019851}
(PI: Trevor Penning), \texttt{R01-LM010098} (PI: Moore), and
\texttt{R01-LM012601} (PI: Moore).

\bibliography{tpotnn}

\appendix

\section{Implementing TPOT-NN}\label{lab:app-impl}
TPOT-NN is implemented as a new feature within the existing TPOT
software, and is available for use in current releases of the
software. TPOT operators consist of 3 main types: feature selectors
(e.g., filtering out features with low variance), feature transformers
(e.g., min-max scaling), and estimators (e.g., any classification or
regression model). Prior to TPOT-NN, all estimators were imported
directly from the two third-party Python libraries Scikit-learn and
XGBoost. We found that existing `plug-and-play' neural network
estimators perform poorly in many case. Furthermore, one of the most
powerful characteristics of neural network models is their extreme
flexibility for defining custom architectures intended to solve
specific analyses.

With this in mind, we implemented TPOT-NN as the following 2
components:

\begin{itemize}
\item An abstract base class (named \texttt{PytorchEstimator}) that
  adds a scikit-learn--like API interface consistent to existing TPOT
  operators to user-defined neural network models implemented using
  PyTorch.
\item Two simple PyTorch neural network models (a logistic regression
  classifier and a multilayer perceptron classifier with 1 hidden
  layer) to serve as benchmarks for TPOT-NN performance, both of which
  are used for evaluation of TPOT-NN in this study.
\end{itemize}

\subsection{Using predefined PyTorch estimators in TPOT-NN}
PyTorch estimators are disabled by default, due to their lengthy
training duration and higher computational requirements. However,
enabling them is done by simply initializing the
\texttt{TPOTClassifier} object using a configuration dictionary
containing the PyTorch estimators.

\subsection{Defining custom PyTorch estimators to use with TPOT-NN}
New TPOT-NN estimators can be added by defining a new Python class
that inherits from the \texttt{PytorchClassifier} mixin class. The new
estimator should implement two components:

\begin{itemize}
\item An instance attribute named \texttt{network} that inherits from
  \texttt{torch.nn.Module}, defining the structure of the neural
  network.
\item A method named \texttt{\_init\_model()} that assigns a number of
  attributes required for all TPOT-NN estimators.
\end{itemize}

To enable the new estimator, the user should provide TPOT with a
configuration dictionary containing the module path for the new
estimator along with any trainable hyperparameters.

For the complete documentation, please refer to TPOT's user guide
(found at \url{http://epistasislab.github.io/tpot/using/} at the time
of writing).

\end{document}